\documentclass[lettersize,journal]{IEEEtran}
\usepackage{amsmath,amsfonts}
\usepackage{algorithmic}
\usepackage{algorithm}
\usepackage{array}
\usepackage{textcomp}
\usepackage{stfloats}
\usepackage{url}
\usepackage{verbatim}
\usepackage{graphicx}
\usepackage{cite}
\usepackage{acronym}
\usepackage{xcolor}

\usepackage{colortbl}

\usepackage{soul}

 \usepackage{capt-of}    
\usepackage{tabularx}
\usepackage{nomencl}
\makenomenclature

\usepackage{subcaption}
\usepackage[T1]{fontenc}

\usepackage{orcidlink}
\usepackage{hyperref}
\hypersetup{
    colorlinks,
    citecolor=black,
    filecolor=black,
    linkcolor=black,
    urlcolor=black
}

\acrodef{AARE}{average absolute relative error}
\acrodef{ANF}{antiresonant nodeless fiber}
\acrodef{ARF}{antiresonant fiber}

\acrodef{CDF}{cumulative distribution function}
\acrodef{CL}{confinement loss}

\acrodef{DNANF}{double-nested antiresonant nodeless fiber}

\acrodef{FNR}{false negative rate}
\acrodef{FPR}{false positive rate}

\acrodef{HCF}{hollow core fiber}

\acrodef{NANF}{nested antiresonant nodeless fiber}
\acrodef{NN}{neural network}
\acrodef{NMSE}{normalized mean square error}

\acrodef{PBGF}{photonic band gap fiber}

\acrodef{ADC}{analog-to-digital converter}
\acrodef{AE}{autoencoder}

\acrodef{ASE}{amplified spontaneous emission}
\acrodef{ASIC}{application-specific integrated circuit}
\acrodef{AUX}{auxiliary}
\acrodef{AWGN}{additive white Gaussian noise}
\acrodef{BER}{bit error rate}
\acrodef{BPD}{balanced photodetector}
\acrodef{BPS}{blind phase search}
\acrodef{CD}{chromatic dispersion}

\acrodef{CDC}{chromatic dispersion compensation}
\acrodef{CELU}{continuous exponential linear unit}
\acrodef{C-FEC}{concatenated FEC}
\acrodef{CMA}{constant modulus algorithm}
\acrodef{CMOS}{complementary metal-oxide semiconductor}
\acrodef{CNN}{convolutional neural networks}
\acrodef{COP}{coefficient of performance}
\acrodef{CPR}{carrier phase recovery}
\acrodef{CPS}{coplanar strip}
\acrodef{CR}{clipping ratio}
\acrodef{CV}{cylindrical vector}
\acrodef{CW}{continuous wave}
\acrodef{DAC}{digital-to-analog convert}
\acrodef{DCI}{data center interconnect}
\acrodef{De-MUX}{demultiplexer}
\acrodef{DFB}{distributed feedback}
\acrodef{DFE}{decision feedback equalizer}
\acrodef{DL}{direct learning}

\acrodef{DNN}{deep neural network}
\acrodef{DPD}{digital pre-distortion}
\acrodef{DP-IQM}{dual-polarization IQ modulator}
\acrodef{DSP}{digital signal processing}
\acrodef{EB2B}{electrical back-to-back}
\acrodef{ECL}{external cavity laser}
\acrodef{EDFA}{erbium-doped fiber amplifiers}
\acrodef{ENoB}{effective number of bits}
\acrodef{ER}{extinction ratio}
\acrodef{ESNR}{electrical signal-to-noise ratio}
\acrodef{EVM}{error vector magnitude}
\acrodef{FEC}{forward error correction}
\acrodef{FEM}{finite element method}
\acrodef{FFNN}{feed forward neural network}
\acrodef{FIR}{finite impulse response}
\acrodef{FMF}{few-mode fiber}
\acrodef{FOC}{frequency offset compensation}
\acrodef{FS}{frequency-selective channel}
\acrodef{GAN}{generative adversarial network}
\acrodef{GRIN}{graded index}
\acrodef{GS}{geometric shaping}
\acrodef{HWP}{half-wave plate}
\acrodef{i.i.d.}{independent and identically distributed}
\acrodef{I/Q}{in-phase and quadrature}
\acrodef{IA}{implementation agreement}
\acrodef{IL}{indirect learning}
\acrodef{ILC}{iterative learning controller}
\acrodef{IMDD}{intensity-modulation direct detection}
\acrodef{ISI}{inter-symbol interference}
\acrodef{LCP}{left circularly polarized}
\acrodef{LP}{linearly polarized}
\acrodef{LSTM}{long short term memory}
\acrodef{LUT}{look up table}
\acrodef{MCF}{multicore fiber}
\acrodef{MMF}{multimode fiber}
\acrodef{MDM}{mode division multiplexing}
\acrodef{MFA}{mode field adapters}
\acrodef{MIMO}{multiple-input multiple-output}
\acrodef{ML}{maximum likelihood}
\acrodef{MMSE}{minimum mean square error}
\acrodef{MRM}{microring modulator}
\acrodef{MSA}{Multi Source Agreement}
\acrodef{MSE}{mean square error}
\acrodef{MUX}{multiplexer}
\acrodef{MZI}{Mach-Zehnder interferomenters}
\acrodef{MZM}{Mach-Zehnder modulator}
\acrodef{OAM}{orbital angular momentum}
\acrodef{OB2B}{optical back-to-back}
\acrodef{OBF}{optical bandpass filter}
\acrodef{OFDM}{orthogonal frequency division multiplexing}
\acrodef{OFDM}{orthogonal frequency division multiplexing}
\acrodef{OIF}{Optical Internetworking Forum}
\acrodef{OOK}{on off Keying}
\acrodef{OPA}{optical phased array}
\acrodef{OSA}{optical spectrum analyzer}
\acrodef{OSNR}{optical signal-to-noise ratio}
\acrodef{PA}{power amplifier}
\acrodef{PAM}{pulse amplitude modulation}
\acrodef{PAPR}{peak-to-average power ratio}
\acrodef{PD}{photodetector}
\acrodef{pdf}{probability density function}
\acrodef{PDM}{polarization-division multiplexing}
\acrodef{PIC}{photonic integrated circuit}
\acrodef{PN}{phase noise}
\acrodef{PN}{phase noise}
\acrodef{PON}{passive optical network}
\acrodef{PRBS}{pseudo random binary sequence}
\acrodef{PS}{probabilistic shaping}
\acrodef{QAM}{quadrature amplitude modulation}
\acrodef{QPSK}{quadrature phase-shift keying}
\acrodef{QWP}{quarter-wave plate}
\acrodef{RC}{raised cosine}
\acrodef{RCF}{ring core fiber}
\acrodef{RCP}{right circularly polarized}
\acrodef{RF}{radio frequency}
\acrodef{RMS}{root mean square}
\acrodef{RNN}{recurrent neural network}
\acrodef{ROSNR}{required optical signal-to-noise ratio}
\acrodef{RTO}{real-time oscilloscope}
\acrodef{RX}{receiver}
\acrodef{SBRNN}{sliding window bidirectional recurrent network}
\acrodef{SOI}{silicon on insulator}
\acrodef{SDM}{space division multiplexing}
\acrodef{SEM}{scanning electron microscope}
\acrodef{SER}{symbol error rate}
\acrodef{SiP}{silicon photonics}
\acrodef{SLM}{spatial light modulator}
\acrodef{SMF}{single mode fiber}
\acrodef{SNR}{signal-to-noise ratio}
\acrodef{SOA}{semiconductor optical amplifier}
\acrodef{SPS}{samples per symbol}
\acrodef{SR}{suppression ratio}
\acrodef{SSMF}{standard single mode fiber}
\acrodef{TEC}{thermoelectric controller}
\acrodef{TIA}{transimpedance amplifier}
\acrodef{TW-MZM}{traveling-wave Mach-Zehnder modulator}
\acrodef{TX}{transmitter}
\acrodef{VNA}{vector network analyzer}
\acrodef{VOA}{variable optical attenuator}
\acrodef{WDM}{wavelength division multiplexing}
\acrodef{XT}{crosstalk}

\begin{document}

\title{Machine Learning Models to Identify Promising Nested Antiresonance Nodeless Fiber Designs}

\author{
Rania~A.~Eltaieb,  Sophie~LaRochelle  and Leslie~A.~Rusch}

\maketitle

\begin{abstract}
Hollow-core fibers offer superior loss and latency characteristics compared to solid-core alternatives, yet the geometric complexity of \acp{NANF} makes traditional optimization computationally prohibitive. We propose a high-efficiency, two-stage machine learning framework designed to identify high-performance \ac{NANF} designs using minimal training data. The model employs a \ac{NN} classifier to filter for single-mode designs (suppression ratio $\ge$~50~dB), followed by a regressor that predicts \ac{CL}. By training on the common logarithm of the loss, the regressor overcomes the challenges of high dynamic range. Using a sparse data set of only 1,819 designs, all with \ac{CL} greater or equal to 1~dB/km, the model successfully identified optimized designs with a confirmed \ac{CL} of 0.25~dB/km. {This demonstrates the \ac{NN} has captured underlying physical behavior and is able to extrapolate to regions of lower \ac{CL}}. We show that small data sets are sufficient for stable, high-accuracy performance prediction, enabling the exploration of design spaces as large as $14e6$ cases at a negligible computational cost compared to finite element methods. 

\color{black}
\end{abstract}
\acresetall


\section{Introduction}

\Acp{HCF} have exceptional performance  compared with standard solid-core fibers; they offer low dispersion, ultra low nonlinearity and 30\% lower latency \cite{poletti2014nested,numkam2023loss}. While solid core fibers rely on total internal reflection for guidance, \acp{HCF} use  photonic bandgap or antiresonance reflections \cite{richardson2017hollow,wei2017negative,poletti2013hollow}. We focus on \acp{ARF} \cite{wei2017negative}, as they have lower loss, wider bandwidth and less complex structures than photonic bandgap fibers \cite{poletti2013hollow}. 

In \acp{ARF} a single layer of hollow capillary tubes in the cladding creates an antiresonant reflection that confines the fundamental mode within the air core. An \ac{ANF} does not allow the tubes to touch one another, eliminating parasitic resonances associated with the glass-to-glass junction. Nesting a small capillary inside each of the main capillaries, \ac{NANF}, greatly enhances performance. A recent breakthrough in \acp{DNANF} \cite{petrovich2025broadband} resulted in loss $<$~0.1~dB/km, lower than standard \acp{SMF}. 

Previously, we used exhaustive search of the geometric parameters for \ac{ANF} to find improved designs \cite{eltaieb2025impact}.  The parameter space for \ac{NANF} or \ac{DNANF} is much larger than \ac{ANF}, and the comprehensive search method is impractical. A more efficient, faster method is required.

Machine learning is a powerful tool in the design and modeling of optical components in general  \cite{schubert2022inverse,kim2025inverse,fernandez2024differentiable}, and optical fibers in particular \cite{he2020machine,meng2023artificial,meng2021use,jewani2024accurate,chugh2019machine}. Fiber models allow for accurate  prediction of performance metrics, and enable fast numerical optimization algorithms. Machine learning models used for optical fiber design include inverse and direct models.

With machine learning inverse design \cite{he2020machine,meng2023artificial}  researchers fix a subset of the fiber design parameters and performance goals. A \ac{NN} is trained to identify the remaining parameters that meet the performance goals. Inverse machine learning has been applied to few mode fibers \cite{he2020machine} and \acp{ARF} \cite{meng2023artificial}.  
These models are very difficult to exploit as a great many solutions may meet the performance goals, i.e., the non-unique solution challenge \cite{kabir2008neural}. 

With machine learning direct models \cite{meng2021use,jewani2024accurate,chugh2019machine}, an \ac{NN} is trained on the totality of fiber design parameters to predict performance. 
In \cite{meng2021use}  the loss of \acp{ARF} with a circular and elliptical capillary structure is predicted by an \ac{NN}.  The \ac{NN} acts as a classifier, mapping a design to a range of \ac{CL}.  Their proposed method  used 
more than 290,000 designs in the data set. A classifier was also used to model a \ac{NANF} structure in  \cite{jewani2024accurate}, with a data set of 60,000 designs. These approaches were very time consuming due to the size of the data set; each design must be run through a finite element solver. Furthermore, using a classifier does not produce accurate predictions of \ac{CL}, but only qualitative metrics.

Using regression \acp{NN} to predict the \ac{CL} is challenging due to the wide range of \ac{CL} values encountered in \ac{HCF} designs, no matter the data set size. This led to the use of classifiers described previously. 
Regression \acp{NN} were used successfully to predict \ac{CL} in photonic crystal fibers (with solid core) \cite{chugh2019machine} and \acp{NANF} in \cite{zhenyu2024confinement}.  The \acp{NN} were trained on the logarithm of the dB per length value of confinement loss, which improved the prediction accuracy. 
However, the single mode guidance is not considered in these models.  We apply the logarithmic regression methodology to explicitly identify single mode \ac{NANF} designs as well as predict \ac{CL}.

Our research focuses on manufacturable fibers with circular geometry defined by a 4-dimensional parameter space. This contrasts with \cite{zhenyu2024confinement} that employed \acp{CNN} to predict \ac{CL} using a structured multi‑dimensional array  of size 16$\times$2$\times$3 to represent  diverse \ac{ARF} structures including \ac{NANF}. In \cite{zhenyu2024confinement}, the extensive data set of 78,312 designs included several samples with ultra-low loss. We demonstrate that a sparse data set of only 1,819 designs with a 6$\times$1 input to the \ac{NN} is sufficient to identify high-performance fibers. Notably, although our training data was limited to a minimum loss of 1~dB/km, our model successfully extrapolated to identify optimized designs with a confirmed CL as low as 0.25~dB/km. Furthermore, our targeted design space facilitates significantly higher predictive precision, yielding a 5\% relative error compared to the 10\% relative error reported for the higher-dimensional model \cite{zhenyu2024confinement}.

By exploiting \acp{NN} to search over an extremely large design space, we achieve performance enhancement at a negligible computational cost compared to finite element methods. We examine prediction accuracy of confinement loss at a central wavelength for various data set sizes and find that even small training sets provide high predictive precision for single-mode \ac{NANF} optimization.

This paper is organized as follows. We present our \ac{NANF} geometry, the initial design space, and our strategy for identifying good designs in Section~\ref{sec:studied_geo}. We describe the \ac{NN} structures and the preparation of the data set of different sizes in Section~\ref{sec:NN_model}. We discuss the accuracy of the converged \ac{NN} models  in Section~\ref{Sec:NN_accuracy}. 
In Section~\ref{sec:search_best_designs}, we discuss  the effectiveness of the  \ac{NN} models  to identify promising designs as a function of data set size and search set size. In Section~\ref{sec:factor_impact} we focus on additional factors influencing the performance improvement that we can achieve with our methodology. In Section~\ref{sec:discussion} we select one optimized design and show simulation results for confinement loss over an extended wavelength range.  Finally, we conclude in Section~\ref{sec:conclusion}.

\begin{figure}[!tb]
\centering
\includegraphics[width=0.4\textwidth]{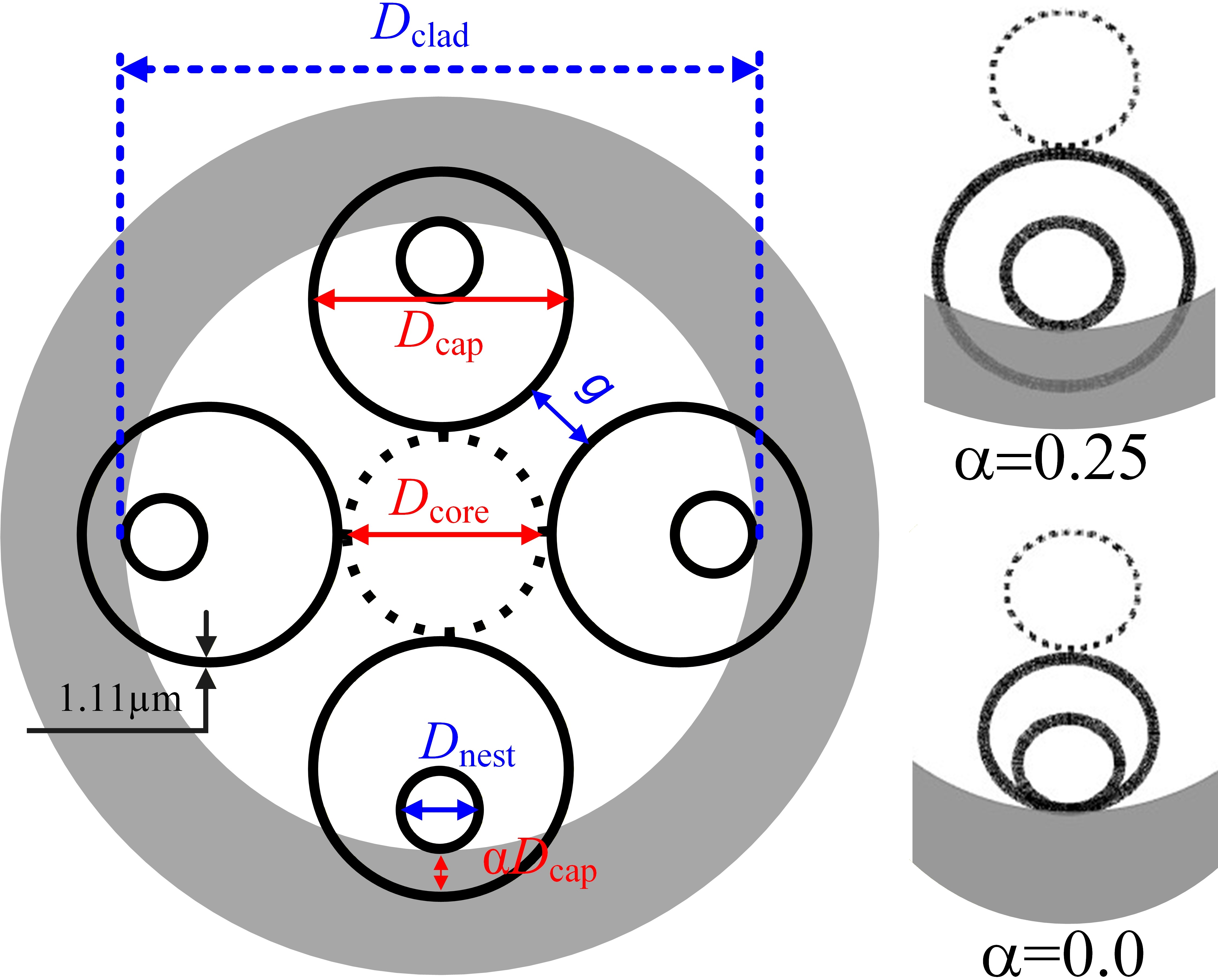}%
\caption{Parameterization of a \ac{NANF} geometry with examples of nested nodes for two $\alpha$.}
\label{fig:geometry}
\end{figure}

\section{Fibers Studied }
\label{sec:studied_geo}

\subsection{Geometry}

On the left of Fig.~\ref{fig:geometry} we sketch a 4-capillary \acf{NANF}; all structures are circular. Black and gray indicate silica, while white indicates air. The inner cladding diameter is $D_{clad}$. The four main capillaries have the same inner diameter $D_{cap}$ with uniform gap separation $g$. The capillaries may penetrate the cladding surface, that is, become embedded in the cladding. We define $\alpha$ as the percentage of the capillary diameter embedded within the cladding.  On the right side of the figure, we demonstrate two examples of $\alpha$. A capillary with an inner diameter $D_{nest}$ is nested inside each main capillary, tangent to the cladding surface (no embedding). The four main capillaries delimit the hollow core diameter $D_{core}$ (dotted circle). All physical dimensions are in $\mu$m.

\subsection{Initial Design Space}
\label{sec:geo_eq}

 We select a capillary thickness of 1.11~$\mu$m for fabrication feasibility and to center the second antiresonance window at 1400~nm (covering the C-band). There remains a vector of four independent parameters fixing a fiber geometry (design): $[ D_{core}  \; D_{cap} \; \alpha \; D_{nest} ]$. The parameters $D_{clad}$ and $g $ are determined by this vector. 

With our definition of $\alpha$, the air space between the nested capillaries is 
\begin{equation}
     \left( 1-\alpha \right) D_{cap}-D_{nest}-1.11\left( 1+2\alpha \right).  
   \label{eq:air_space}
\end{equation}

Good fiber designs have  modes in the main and nested capillaries phase matched. If $D_{nest}$ is too large, the fundamental mode will be lost from the core area and the suppression of all modes will be high. We reject

\begin{equation}
    D_{nest}\leq\left( 1-\alpha \right) D_{cap}-D_{core}.
    \label{equ:condition_on_fund}
\end{equation}
\noindent For reasonable fabrication, we further require $D_{nest}$ to be greater than 0.1$(1-\alpha) D_{cap}$~$\mu$m.

Our design space is delimited by 
\begin{equation}
\begin{aligned}
      D_{core}=&~[20.0\colon1.0~\colon 60.0],\\
      D_{cap}=&~[25.8\colon0.5~\colon54.3]\\ 
    \alpha=&~[0.0~\colon0.05\colon 0.5]\\
    D_{nest}=&~[0.1~\colon0.05\colon0.6]*(1-\alpha)D_{cap}
\end{aligned}
\label{eq:Used_geo_train}
\end{equation}

\noindent We find the inner cladding diameter $D_{clad}$ for each design per
\begin{equation}
    D_{clad}=D_{core}+2(1-\alpha)(D_{cap}+2.22)
    \label{equ:Dclad}
\end{equation}
 \noindent and the gap $g$  per
  \begin{equation}
    g =sin\left(\frac{\pi}{4}\right)D_{core}+\left(sin\left(\frac{\pi}{4}\right)-1\right)\left(D_{cap}+2.22\right)
    \label{equ:gap}
\end{equation}
\noindent We reject designs with  
\begin{equation}
    g <3 \And{g >6}.
    \label{equ:condition_on_gap}
\end{equation}
\noindent For $g<$3~$\mu$m  there would be high resonance loss; for  $g>$6~$\mu$m there would be high leakage loss. A final check rejects designs where the nested capillary grows large enough to touch the main capillary near the core.

\subsection{Ground Truth Data Set via COMSOL}
\label{Sec:Training set preparation}

We identified 18,422 \ac{NANF} in the design space that respected the conditions in (\ref{equ:condition_on_fund}) and (\ref{equ:condition_on_gap}). We used COMSOL to evaluate each design at the center of the antiresonance window studied ({1400~nm}). From this simulation, we found the confinement loss in dB/km of the fundamental mode ($CL$)  and the first higher order mode ($CL^{\text{1st}}$).  We define the suppression ratio, $SR$,  by
\begin{equation}
    SR~[\mathrm{dB}] = CL^\text{1st}~[\mathrm{dB/km}] - CL~[\mathrm{dB/km}]
    \label{eq: SR}
\end{equation}
\noindent An $SR\geq$~50~dB is a good predictor of single mode behavior.

The lowest \ac{CL} produced by COMSOL in our data set is 0.276~dB/km.
Our goal is to examine the ability of a \ac{NN} to identify very low loss designs \textbf{without having seen those designs in the data set}. For this reason, we excluded 234 designs with $CL$ less than 1~dB/km from the ground truth data set of 18,422 \acp{NANF} designs. We reserve these good designs to compare with the \ac{NN} improved designs in our final remarks.

\subsection{Modeling Strategy}
\begin{figure}[t]
    \centering
    \includegraphics[width=0.9\linewidth]{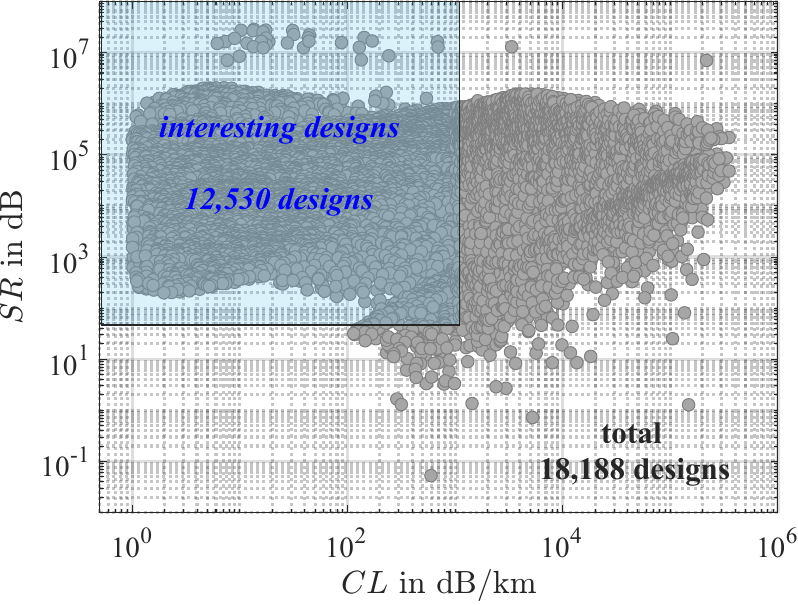}
    \caption{For the initial design space  of 18,188 designs: suppression ratio vs. fundamental mode confinement loss at 1400~nm.}
    \label{fig:CLvsSR}
\end{figure}

We present the scatterplot of confinement loss $CL$ and suppression ratio $SR$ in Fig.~\ref{fig:CLvsSR} for the data  set of 18,188 designs. We observe a large dynamic range for both performance metrics. This poses significant challenges for \ac{NN} models used to predict performance. We can address this issue by adjusting our goals. 

The highlighted region in Fig.~\ref{fig:CLvsSR} has 12,530 designs that we label as \textit{interesting}. An interesting design is single mode, which we define as $SR\geq$~50~dB. Furthermore, to be interesting the $CL$ must be less than $10^3$~dB/km. Our strategy will be to use a \ac{NN} classifier to identify that a design is interesting and a \ac{NN} regression to estimate the $CL$ of the design. In this way, the \ac{NN} regression faces a much more reasonable dynamic range for performance predication. 

The \ac{NN} model for \ac{NANF} performance will run orders of magnitude faster than COMSOL and permit a fine search of \ac{NANF} parameter space. While the classifier passes \ac{CL} as high as $10^3$~dB/km, we will assess regressor performance on designs with \ac{CL} below 6.3~dB/km. Error in \ac{CL} prediction above this threshold will not impact the ability to identify a good fiber design (i.e., one at low \ac{CL}) when searching the  \ac{NANF} parameter space with the \ac{NN}.

\color{black}

\color{black}

\section{Neural Network Model}
\label{sec:NN_model}

\begin{figure}[h!]
    \centering
    \includegraphics[width=1\linewidth]{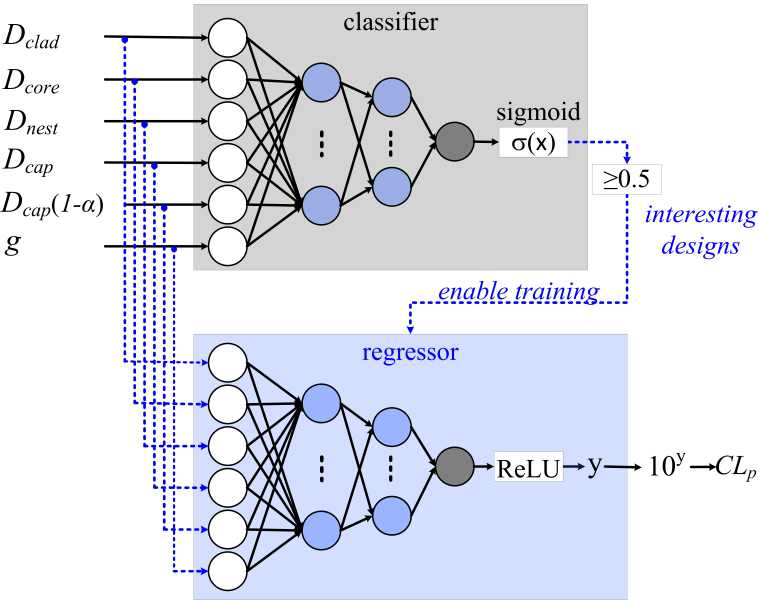}
    \caption{Classifier and regressor neural network structures for training/testing with dataset of $n$ designs.}
    \label{fig:NN_architecture}
\end{figure}

In Fig.~\ref{fig:NN_architecture} we show our structure for modeling \ac{NANF} performance with the goal of identifying fibers with improved performance.  \textit{Interesting} designs are identified via a classifier \ac{NN}. The classifier weights are fixed and designs with a positive classification (interesting) are sent to train a regression \ac{NN} that estimates the confinement loss of the fundamental mode.  The 6-dimensional feature  $D=[ D_{clad} \;D_{core}  \; D_{nest} \; D_{cap} \; D_{cap}(1-\alpha) \;g ]$  is a single design and is input to both \acp{NN}.

\subsection{Data set preparation}
The master data set of size $n_{max}$=18,188 designs may be larger than is needed for creation of a good \ac{NN} predictor of \ac{NANF} performance. As the simulation time to create the ground truth data is high, we will examine the complexity/accuracy trade-off in data set size. We will set data set size $n$ and randomly select a subset of that size from the master set. Multiple random selections will sometimes be made for a given $n$.

For each data set of size $n$, we will randomly partition the set into a training set with 80\% of the designs, a validation set with 10\% of the designs, and a test set with the remaining 10\%. Multiple random partitions will sometimes be made.

\subsection{Classification NN for Identifying Interesting Designs}

In Fig.~\ref{fig:NN_architecture} the first stage is a fully connected feedforward \ac{NN} classifier. The \ac{NN} classifier has two hidden layers, the first with 70 neurons and the second with 50 neurons. The output layer has one neuron for binary classification to identify an interesting design.

The neuron weights are initialized randomly. We evaluated several activation functions for the hidden layers, and opted for the ReLU function as it offered faster convergence and better performance. We selected the Adam optimizer after evaluating several options.

We use the sigmoid function for the cross-entropy loss function,
\begin{equation}
 \sigma(x) = \frac{1}{1 + e^{-x}}.
\end{equation}
As we have two classes, we threshold the output to achieve the binary classification. The best learning rate we identified was 0.001 for 5000 epochs. A summary of {the classifier} hyper-parameters is provided in the first column of Table~\ref{tab:hyperparameters}. 
Once classifier training is complete, we fix the weights and begin regression training. 

\subsection{Regression NN for Confinement Loss Prediction}

The regressor \ac{NN} illustrated in Fig.~\ref{fig:NN_architecture} is a fully connected feedforward \ac{NN} predicting confinement loss. The entire data set is input to the regression \ac{NN}, however, weights are only adapted for interesting designs. Inherently, the regression \ac{NN} sees less data than the classifier \ac{NN}.  Approximately 70\% of the designs examined are classified as interesting designs. This percentage depends on our criteria for maximal $CL$ to be considered \textit{interesting}.

The regressor has two hidden layers, each with a distinct numbers of neurons. In Table~\ref{tab:hyperparameters} we summarize \ac{NN} hyperparameters for the classifier and regressors. Data set size had an impact on the best choice for layer size and other hyperparameters. 
 Regressor~1 hyperparameters were used for larger data sets, $n>9,000$, and  regressor~2 hyperparameters were used for  smaller data sets,  $n\leq 9,000$.
We use the ReLU nonlinear activation function to accelerate convergence and enhance performance. We initialize the weights randomly and use the Adam algorithm for faster convergence and improved performance.

We define the true confinement loss as $CL_t$ and the predicted value from the \ac{NN} regressor as $CL_p$. Despite reducing the dynamic range of $CL_t$ to $10^3$ in the data set, convergence in training is still difficult~\cite{chugh2019machine}. We therefore predict the common logarithm of the loss, $y$, as seen in Fig.~\ref{fig:NN_architecture}. The loss is in dB/km, so this is essentially a double logarithm. We convert $y$ to dB/km and report predictions as $CL_p$ in dB/km. 

We studied multiple loss functions, and the \ac{NMSE} loss function showed faster \ac{NN} regressor convergence.  We first estimate the \ac{MSE} for the training 
\begin{equation}
        MSE(n)=\frac{1}{0.8n}\sum_{k=1}^{0.8n}\left(\log_{10}(CL_{t,k})-y_k\right)^2
    \label{eq:MSE}
\end{equation}
\noindent where 0.8$n$ reflects training over 80\% of a data set of size $n$.
We define mean
\begin{equation}
    \mu_t=\frac{1}{0.8n}\sum_{k=1}^{0.8n}\left(\log_{10}(CL_{t,k})\right)
\end{equation}
\color{black}
\noindent and variance \begin{equation}
        \sigma_t^2(n)=\frac{1}{0.8n}\sum_{k=1}^{0.8n}\left(\log_{10}(CL_{t,k})-\mu_t\right)^2.
    \label{eq:variance_CLtrue}
\end{equation}
\noindent The \ac{NMSE} is therefore as
\begin{equation}
    NMSE(n)=\frac{MSE(n)}{\sigma_t^2(n)}.
\end{equation}

\begin{table}[t]
    \caption{Hyper-parameters of the classifier and regressors}    
    \footnotesize
      \label{tab:hyperparameters}
      \centering
    \begin{tabular}{|c||c|c|c|}
    \hline
         & Classifier & Regressor~1 & Regressor~2  \\ 
         \hline 
         \# of hidden layers & 2 & 2 &  2\\  
         \hline
         \begin{tabular}{@{}c@{}}\hspace{-1em}\# of neurons in \\  hidden layers\end{tabular}&  \begin{tabular}{@{}c@{}} {70, 50}\end{tabular} & \begin{tabular}{@{}c@{}}  {128, 32}\end{tabular}  &  \begin{tabular} {@{}c@{}}{130, 38}\end{tabular}  \\  
         \hline        
        \hspace{-2.1em}learning rate & 0.001 & 0.001  &  0.0001 \\  \hline
        \hspace{-4.8em}epoch & 5000 & 5000  &  10000 \\  \hline
    \end{tabular}
\end{table}

\color{black}
\section{Neural Network Accuracy}
\label{Sec:NN_accuracy}

In this section we explore the impact of data set size $n$ on \ac{NN} accuracy in prediction of performance. For data sets with $n\leq$1,500, our \ac{NN} overfitted the data or did not converge. We set the smallest data set to examine at $n$=1,819, with eight data set sizes examined in total.

\subsection{Classification Error for Various Data Set Sizes}

We define a trial to be a subset of the master data set and a training/validation/testing partition of that subset. For  $n\!=\!n_{max}$, the data set is always the master data set; we have ten trials, each a different partition of training/validation/testing. For all other $n$, the ten trials take the form of ten distinct subsets randomly chosen from the master data set (each with a single partition into training/validation/testing). 

For a trial $i$, the number of true positives is ($TP_i$); a correctly identified \textit{interesting} design is a true positive. The number of false negatives is ($FN_i$); an \textit{interesting} design that is incorrectly classified is a false negative. The number of false positives is ($FP_i$); an unwanted (not interesting) design classified as interesting is a false positive. The number of  true negatives is ($TN_i$); an unwanted design classified as unwanted is a true negative. These four quantities form the confusion matrix for trial $i$.  
For each data set size $n$, we estimate the \acf{FNR} and \acf{FPR} by averaging over the ten trials
\begin{equation}
\begin{aligned}
FNR(n)=\frac{1}{10}\sum_{j=1}^{10}\frac{FN_j}{FN_j + TP_j}\\
FPR(n)=\frac{1}{10}\sum_{j=1}^{10}\frac{FP_j}{FP_j + TN_j} 
\end{aligned}
\label{eq:FNR_FPR}
\end{equation}

\color{black}

\begin{figure}[!t]
\centering
\includegraphics[width=0.42\textwidth]{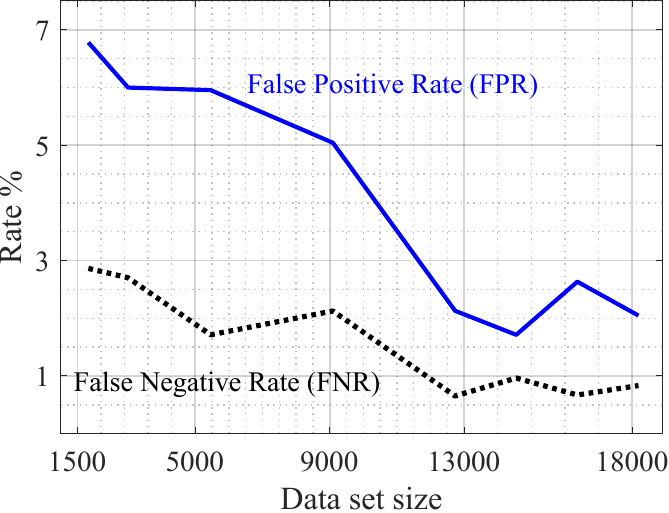}
\caption{{Classifier false positive  and false negative rates vs. data set size.}}
\label{fig:Perf_classifier}
\end{figure}

In Fig.~\ref{fig:Perf_classifier} we plot the \ac{FNR}, dashed curve, and \ac{FPR}, solid curve, rates versus the data set size $n$. As expected, performance improves (false rates decline) as data set size increases. 
This improvement saturates from $n$=12,732 onward. Performance  at  smaller data set size is, nonetheless, acceptable. Even at the smallest $n$, we identify a minimum of about 97.1\% of \textit{interesting designs}. Therefore, we are unlikely to let a good design pass unexamined. The false positive rate is higher (7\% rather than 3\%), but this will only lead to examining some cases unnecessarily (more computation). Hence, for classification, small data set size is sufficient.

\subsection{Regression Error for Various Data Set Sizes}

We quantify regression performance only for designs with relatively low loss. The \ac{NN} will be exploited to identify good designs, hence its performance for designs with $CL_t\leq$~6.3~dB/km is most relevant. The choice of $6.3$~dB/km is ad hoc, but allows for easier visualization and interpretation of results. The training of \ac{NN} weights covered all designs, affecting convergence. Future work could examine including emphasis on lower loss designs in the training process. 

We define absolute relative error $\xi$ as
\begin{equation}
    \xi=\left|\frac{CL_{t}-CL_{p} }{CL_{t}}\right|.
    \label{eq:relative_error}
\end{equation}
\noindent  
For a data set of size $n$, let $S_{n}$ be the set of designs  with $CL_t\leq$~6.3~dB/km collected from the test partition of a trial (trial is as described in the previous section). The average absolute relative error of a given trial, $\eta$, is 

\begin{equation}
   \eta= \frac{1}{|S_n|}\sum_{i=1}^{|S_n|} \xi_i
    \label{eq:AARE_meanstep}
\end{equation}
where $|S_n|$ is the number of elements in $S_n$ and $\xi_i$ is the absolute relative error of the $i^{\text{th}}$ element of $S_n$.
We define the mean of $\xi$ over ten trials
\begin{equation}
   \text{Mean} \; \xi= \frac{1}{10}\sum_{j=1}^{10} \eta_j
    \label{eq:AARE}
\end{equation}

\begin{figure}[!t]
\centering
\subfloat[]{\includegraphics[width=0.44\textwidth]{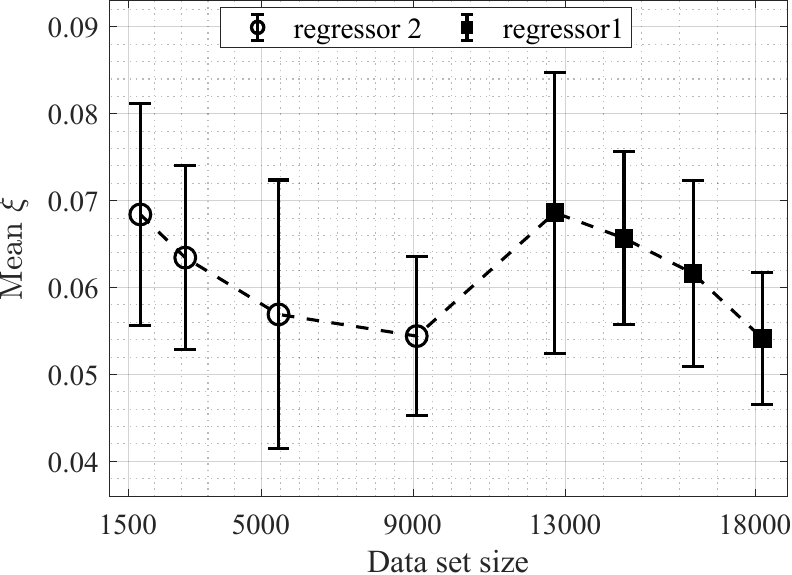}}
\hfil
\subfloat[]{\includegraphics[width=0.42\textwidth]{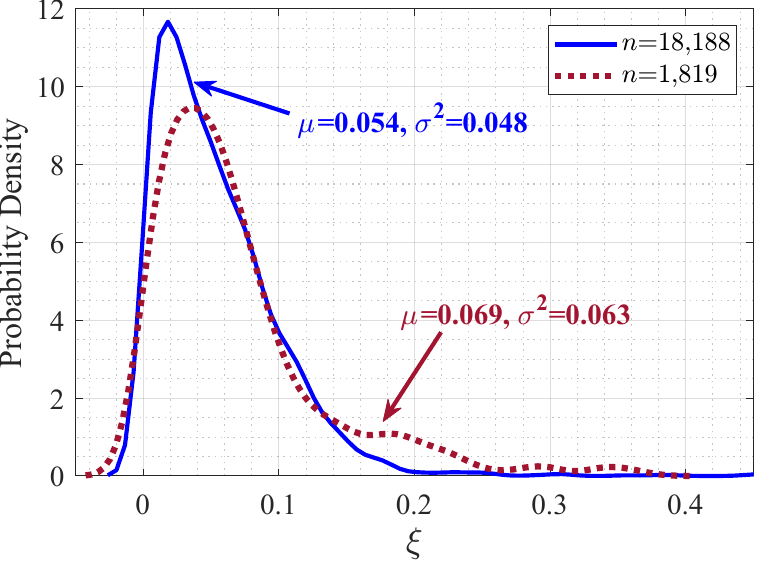}}
\caption{a)~Mean absolute relative error vs. data set size $n$ for designs with $CL\leq$~6.3~dB/km with error bars indicating standard deviation; b)~estimated probability density function of the absolute relative error for $n$=1,819 and $n$=18,188.}
\label{fig:regressor_error}
\end{figure}

In Fig.~\ref{fig:regressor_error}a we plot the estimated mean of average absolute relative error versus the data set size $n$.  The markers indicate the mean values and the error bars represent the standard deviation. Regressor~1 used hyperparameters optimized for larger data set sizes, and regressor~2 used those best for smaller sizes.  Both regressors have lower error as data set size grows. In all cases, error is low, never exceeding 0.07 in the mean for designs with $CL\le$~6.3~dB/km.

We estimated the \ac{pdf} of the absolute relative error $\xi$ using the kernel method on the normalized histograms from all ten trials \cite{hastie2005elements}. In Fig.~\ref{fig:regressor_error}b, we present the estimated \ac{pdf} for data sets of size $n$=18,188 and $n$=1,819. The larger data set size has a shorter tail, so better error performance. The means of 0.069 at $n$=1,819 and 0.054 at  $n$=18,188 reflect this difference. However, the \acp{pdf} are rather similar despite the large difference in data set size. 
\color{black}

\section{Identifying Best Designs}
\label{sec:search_best_designs}

The \ac{NN} can estimate fiber performance at speeds orders of magnitude greater than what could be accomplished with finite element simulations. In this section, we address the main purpose of the trained \ac{NN}: searching over an immense design space, identifying a few most promising designs, and then confirming the performance improvement in a finite element simulation. 
Previous sections reported error over multiple trials of each data set size. Given the time required for confirmation of performance, in this section we restrict our attention to a single trial at each data set size. 

\subsection{Search Space}

We use our trained \ac{NN} model to explore a space of $N \gg n$ designs, i.e., a much larger and denser space than the data set. We randomly sample the design space with a uniform distribution for $D_{core}$ and $D_{cap}$ parameters in the following ranges:

\begin{equation}
\begin{aligned}
      D_{core}&\in~[20.0\colon31.0]\\
    D_{cap}&\in~[25.8\colon54.3]
\end{aligned}
\label{eq:Used_geo_search1}
\end{equation}

\noindent For each random pair $(D_{core},D_{cap})$ we examine a grid of 121 pairs $(\alpha,D_{nest})$  taken from  
\begin{equation}
\begin{aligned}
       \alpha=&[0.0\colon0.02\colon0.5]\\
       D_{nest}=&[0.1\colon0.02\colon0.6]*(1-\alpha)D_{cap}
\end{aligned}
\label{eq:Used_geo_search2}
\end{equation}

\noindent We discard those designs that do not meet the criteria (\ref{equ:condition_on_fund}) and (\ref{equ:condition_on_gap}). We continue to sample $(D_{core},D_{cap})$ until $N \sim14e6$ valid designs have been identified and the loss predicted via \ac{NN}.

\subsection{Performance Improvement with Large Data Set}

\begin{figure}[!t]
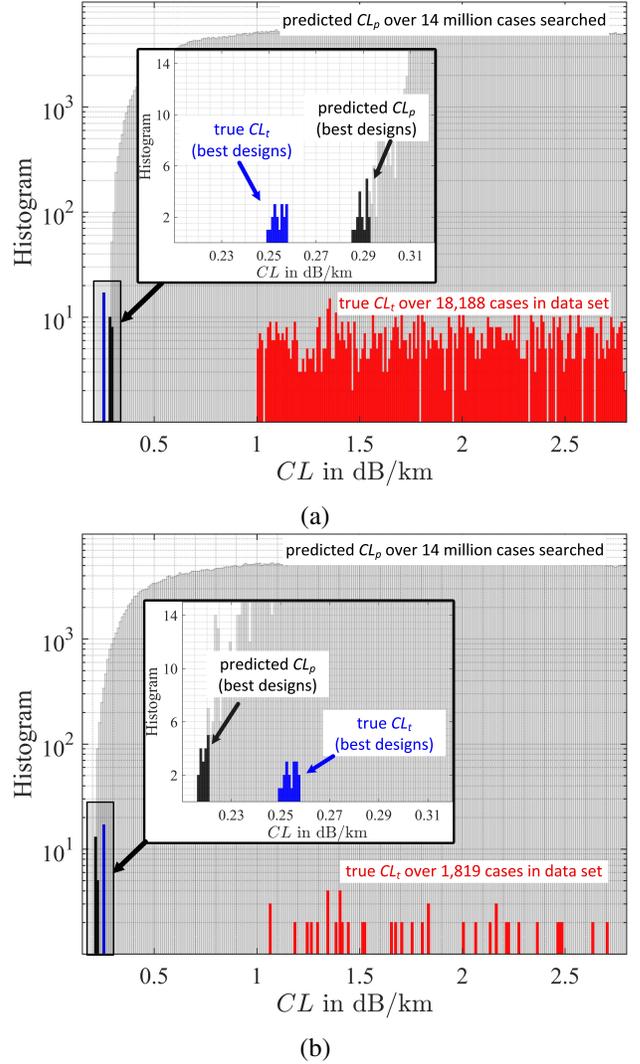

\centering
\subfloat[]{\includegraphics[width=0.45\textwidth]{Figures/Hist_vs_CL_nmax_ver1a.jpg}}
\hfil
\subfloat[]{\includegraphics[width=0.45\textwidth]{Figures/Hist_vs_CL_nmin_ver1a.jpg}}
\caption{ Histograms of: gray predicted $CL_p$ over search space, red true $CL_t$ over data set; inset has predicted $CL_p$ in bold black, and their confirmed, true $CL_t$ in bold blue, for data sets of size a)~$n$=18,188 and b)~$n$=1,819. }
\label{fig:Hist_CL_nmin_nmax}
\end{figure}

We first examine a \ac{NN} trained with the master data set. The search space of $14e6$ designs yields about $1.2e6$ designs with  $CL_p\leq$~2.8~dB/km. The histogram of $CL_p$ for these designs is given in Fig.~\ref{fig:Hist_CL_nmin_nmax}a in gray. The histogram of $CL_t$ from the data set is given in red; 1,139 out of 18,188 designs have $CL_t\leq$~2.8~dB/km.

Consider the far left edge of the gray histogram given in the inset of Fig.~\ref{fig:Hist_CL_nmin_nmax}a. Our \ac{NN} found 62 designs with $CL_p\leq$~0.3~dB/km. We have put in bold black lines the $CL_p$ for the 18 best designs identified by the \ac{NN}. We ran COMSOL on these designs to determine $CL_t$, displayed in bold blue lines in the inset. We limited our examination to 18 designs given the COMSOL simulation time.  

Although the master data set had a minimum $CL_t$ of 1~dB/km, searching over a large search space with an \ac{NN} identified thousands of \ac{NANF} designs with potentially improved performance. The potential was confirmed with finite element simulations.  With this particular trial, the confirmed $CL_t$ is less than the predicted $CL_p$. We confirmed a design with $CL_t\approx$~0.25~dB/km, greatly improving over the data set at 1~dB/km.

\subsection{Performance Improvement using Small Data Set}
We next examine a \ac{NN} trained with the smallest subset of the master data set. Recall that data sets with $n\leq$~1,500 did not converge or overfitted, hence $n$=1,819 is the smallest data set examined. One subset was selected randomly from the master data set, and the results are presented here. 

In Fig.~\ref{fig:Hist_CL_nmin_nmax}b we plot histograms for predicted and true \ac{CL}, as we did in Fig.~\ref{fig:Hist_CL_nmin_nmax}a for the master data set. The gray histogram has the same density for the two figures as the same search area is covered (size $14e6$). The red histogram is much less dense as only 10\% of the master data set is included; for $CL<$~2.6~dB/km there are  126 designs for the small set vs. 1,139 for the large set. In both cases the data set has by construction a minimum $CL_t\approx$~1~dB/km.

The trained \ac{NN} was able  to identify 3646 designs with predicted $CL_p\leq$~0.3~dB/km, as seen in gray in the inset. The lowest $CL_p$ is 0.22~dB/km, with the best 18 designs in black on the inset. The confirmed $CL_t$ of these designs (in blue) show a minimum $\approx$~0.25~dB/km. 
For these two trials, one at $n_{min}$ and one at $n_{max}$, the confirmed, minimum $CL_t$ are similar. Therefore, it appears that a small data set is sufficient  to train a \ac{NN} to find designs with promising performance at low computational cost. It is interesting that for these particular trials, predictions are optimistic for $n_{max}$ and pessimistic for $n_{min}$.

\color{black}

\subsection{Impact of Data Set Size}
\label{sec:n_impact_TS_size}

Our examination in the previous section of a small and large data set size showed that the best designs identified in each case led to similar confirmed performance. We continue to examine this behavior over more data set sizes $n$.

Let $D_i$ be any design in the search space of $M\approx14e6$ designs. We index the designs according to their predicted performance $CL_p(D_i)$. Let $D_1$ be the best design identified by a given \ac{NN}, i.e., $CL_p(D_1)\le CL_p(D_2) \le ... \le CL_p(D_M)$. We examined eight \acp{NN}, each at a distinct $n$ and each trained on one trial, i.e., one random subset of the master data set with a random partition into training/validation/testing. 

\begin{figure}[!t]
\centering
\includegraphics[width=0.47\textwidth]{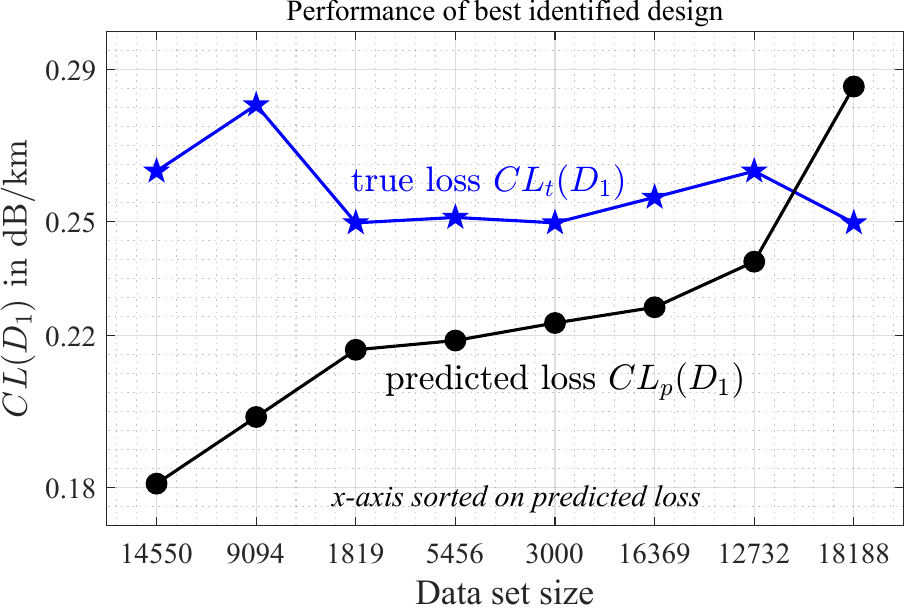}
\caption{Confinement loss of the best designs at each data set size; predicted is $CL_p(D_1)$ and confirmed is $CL_t(D_1)$.}
\label{fig:Allset_CL_results}
\end{figure}

Before plotting, we sorted the \ac{NN} results along $CL_p(D_1)$. The \ac{NN} with $n$=14,550 had the design with the lowest minimum predicted loss, and \ac{NN} with $n$=18,188 had the design with the highest minimum predicted loss. The results are plotted in Fig.~\ref{fig:Allset_CL_results} in the curve with circle markers; the $x$-axis specifies $n$. 

The predicted loss varies from 0.18 to 0.29~dB/km for the data sets examined. 
There is no observable correlation between predicted  $CL_p(D_1)$ and data set size. Two large data sets yield the lowest and highest predicted  \ac{CL}, while the smallest data set provides a median value for predicted  \ac{CL}. We also plot in star markers the confirmed loss $CL_t(D_1)$. The variation of confirmed best-case loss, $CL_t(D_1)$, varies from 0.25 to 0.29~dB/km. While the span of the predicted  $CL_p(D_1)$  is 0.11~dB/km, the confirmed $CL_t(D_1)$ has a span of only 0.03~dB/km. Although most predictions were optimistic, the true \ac{CL} was lower than the predicted  \ac{CL}  at $n$=18,188.

Recall the data sets had \ac{CL} greater than 1~dB/km in all cases. At all data set sizes we see a marked reduction in $CL_t(D_1)$. Therefore, \ac{NN} are very effective in taking a set of fiber designs with mediocre performance and finding greatly improved designs. The \ac{NN} approach has computational cost orders of magnitude lower than finite-element methods. 

From Fig.~\ref{fig:Allset_CL_results}, we conclude that larger size is not needed to identify designs with greatly improved true $CL_t$. Furthermore, the error in prediction does not impact the quality of the design. That is, cases where predicted $CL_p$ was overly optimistic still yielded a design with good true $CL_t$ (0.26~dB/km); maximum error of 0.08~dB/km at data set size of 14,550 still maintains low confirmed loss. The \ac{NN} with the smallest error (\ac{NN} with data set size 12,732)  yields a design with the same true $CL_t$ of 0.26~dB/km. Error is critical in training to get a good \ac{NN}, but less important when considering the best designs found searching over a very dense design space.

\section{Factors Impacting Performance Improvement}
\label{sec:factor_impact}

In section \ref{sec:search_best_designs}, for one trial per $n$, we examined the \acp{NN} performance for 1)~a search set of $14e6$ designs, and 2)~examination of $CL_t(D_1)$. In section \ref{Sec:NN_accuracy} we included multiple trials to examine \ac{NN} error performance. In this section we investigate the impact of these choices.

\color{black}

\begin{figure}[!t]
\centering
{\includegraphics[width=0.45\textwidth]{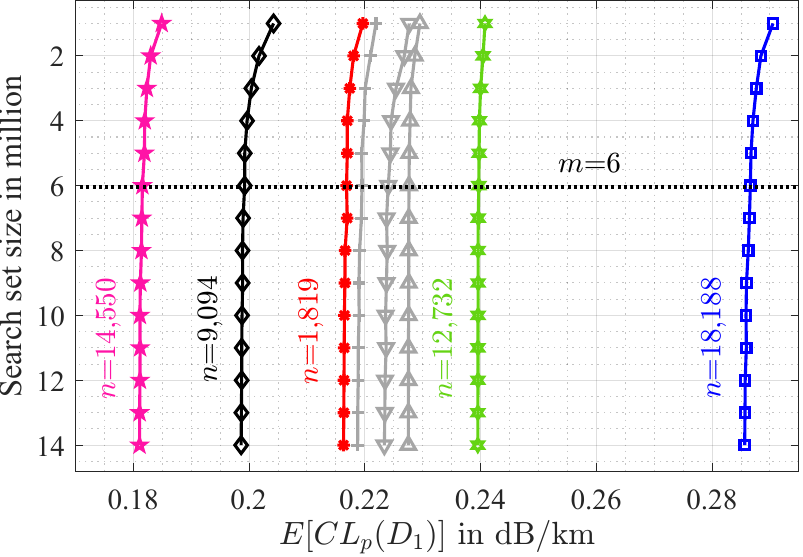}}
\caption{For each data set size $n$, search set size in millions versus  $E[CL_p(D_1)]$ as estimated over 50 trials.}
\label{fig:Nsearchimpact_results}
\end{figure}

\subsection{Impact of search set size on NN best designs}

In Fig.~\ref{fig:Allset_CL_results}, for each $n$, we plot $CL_p(D_1)$ that is the minimum predicted \ac{CL} over the entire search area $A$ with $|A|=14e6$. In this section, with our compiled data over $A$, we consider what might have happened had we used a smaller search area. Specifically, we examine how $CL_p(D_1)$ would be impacted.  We fix a search size $m\le M$ and randomly select 50 subsets of that size. Each subset will have a new best design, with a new best predicted \ac{CL}. We estimate $E[CL_p(D_1)]$ for each $m$ by averaging over the 50 subsets' best predicted \ac{CL}.

Figure~\ref{fig:Nsearchimpact_results} shows the search set size $m$ versus $E[CL_p(D_1)]$. Each curve refers to a data set of size $n$ for training the \ac{NN}.
The ``bottom'' of each curve refers to $CL_p(D_1)$ at $m=M$, i.e., the results in Fig.~\ref{fig:Allset_CL_results}. 

The curves are very steep, showing that our search set size was  much larger than necessary. 
A  search set of $m=6e6$ is sufficient to identify roughly the same minimal predicted \ac{CL}. However, the difference in \ac{NN} execution time for $m=6e6$ or $m=M=14e6$ is negligible. Therefore, we continued our studies with the $14e6$ search set.

\color{black}

\begin{figure}[!t]
\centering
\includegraphics[width=0.47\textwidth]{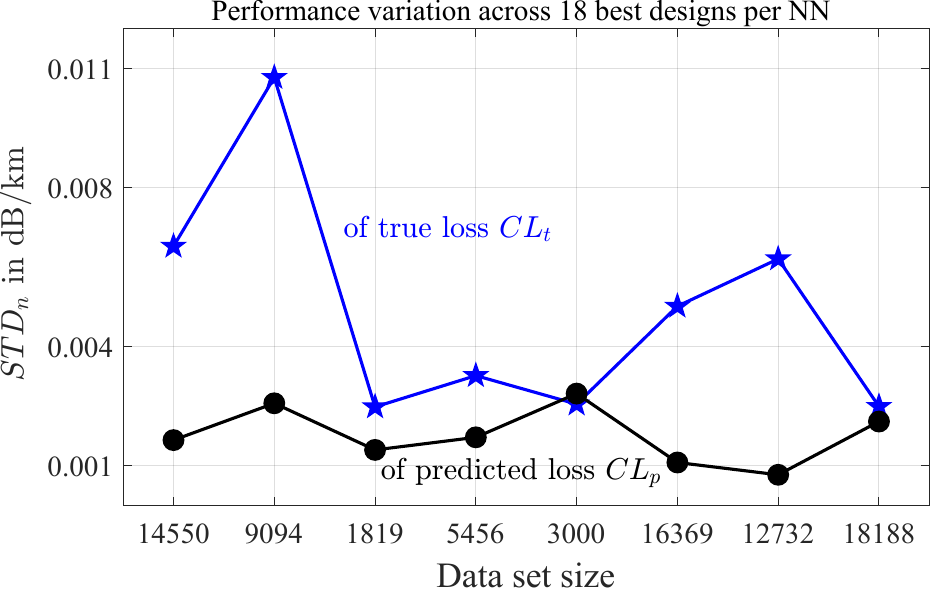}
\caption{Standard deviation of the best 18 predicted and confirmed designs at all studied data set sizes.}
\label{fig:Allset_STD_results}
\end{figure}

\subsection{Relevance of Design with Lowest Predicted Loss}

In the inset of Fig.~\ref{fig:Hist_CL_nmin_nmax}, we highlighted the performance of a subset of the best designs, both predicted and true. As the confirmed true $CL_t$ required finite element simulation, we limited this group to the 18 best designs. In this section, we examine the variation in  predicted and confirmed loss among this group.

For a given \ac{NN} with a data set size $n$, we calculate the standard deviation $STD_{n}$ by 
\begin{equation}
    STD_{n}=\sqrt{\frac{1}{17}\sum_{i=1}^{18}(x_i-\bar{x})^2}
\end{equation}

\noindent where $x_i$ is the confinement loss of design $i$, and $\bar{x}$ is the mean of the confinement loss of the 18 designs. 
In Fig.~\ref{fig:Allset_STD_results}, we plot the standard deviation $STD_{n}$ versus data set size for $CL_t$ and $CL_p$. The $x$-axis has the same ordering as in Fig.~\ref{fig:Allset_CL_results}. 

The standard deviation of the predicted and confirmed losses is small, with a maximum value of 0.011~dB/km. In most cases, the standard deviation of the confirmed loss is higher than that of the predicted loss; there is an exception at $n$=3,000.

\begin{figure}[!t]
\centering
{\includegraphics[width=0.47\textwidth]{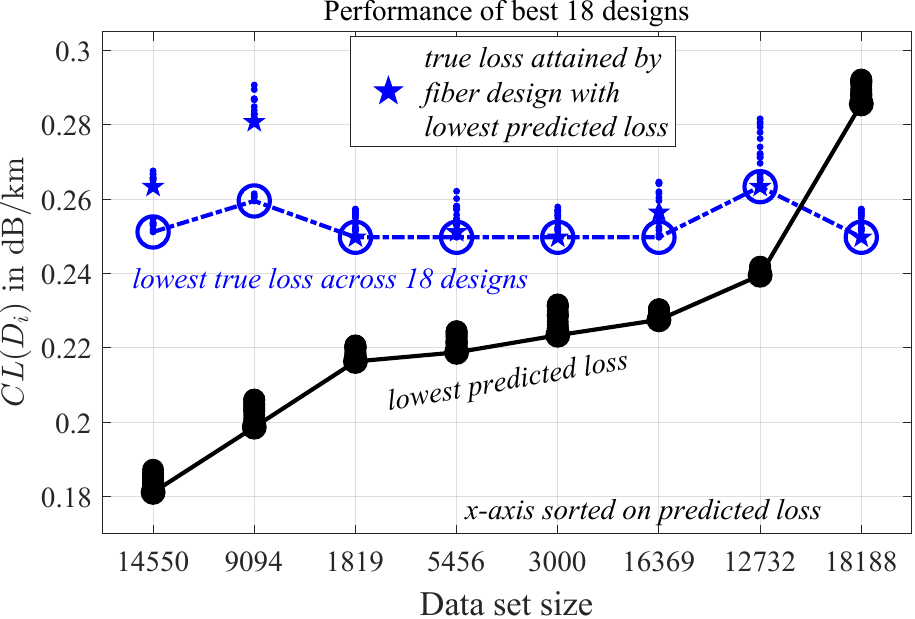}}
\caption{The best predicted confinement loss $CL_p$ in black dots, their confirmed confinement loss $CL_t$, and the best confirmed confinement loss among the 18 design validated in COMSOL versus data set size. }
\label{fig:summary_18_all_n}
\end{figure}

In Fig.~\ref{fig:summary_18_all_n}, we recreate Fig.~\ref{fig:Allset_CL_results} and add information on $CL_t(D_i), \: i\in(2,18)$. The star markers are $CL_t(D_1)$ whereas the empty circle markers are $\min_i CL_t(D_i)$. All other $CL_t(D_i)$ are dots. The $CL_p(D_i)$ are plotted as filled circles. 

The difference in dB/km between star and empty circle markers indicates the improvement in true lowest \ac{CL} if we use finite element to examine a collection of 18 most promising designs. The difference is small, but could prove worthwhile. Our proposed methodology is to use a few minutes to search with \ac{NN} and a few hours (in the case of 18 cases) to refine with finite element. Training time of the \ac{NN} is negligible (included in the few minutes).

\subsection{Impact of Data Set Realization}

In section~\ref{sec:search_best_designs}, we saw in Fig.~\ref{fig:Allset_CL_results} that the smallest data set could produce good \ac{NANF} designs. These results were based on a single trial at each data set size $n$. In this subsection, we investigate multiple trials for $n=n_{min}$ before generalizing our conclusion.

We randomly select 20 subsets of size $n$=1,819. Each subset is divided into a training/validation/testing partition to train an \ac{NN}.  Based on the observations in the previous subsections, we fix the search set of size $M=14e6$, and we begin by examining only the $D_1$ design output in each trial. Finally, we examine the statistics of performance of the 18 best designs over all trials. 

\begin{figure}[!t]
\centering
\includegraphics[width=0.45\textwidth]{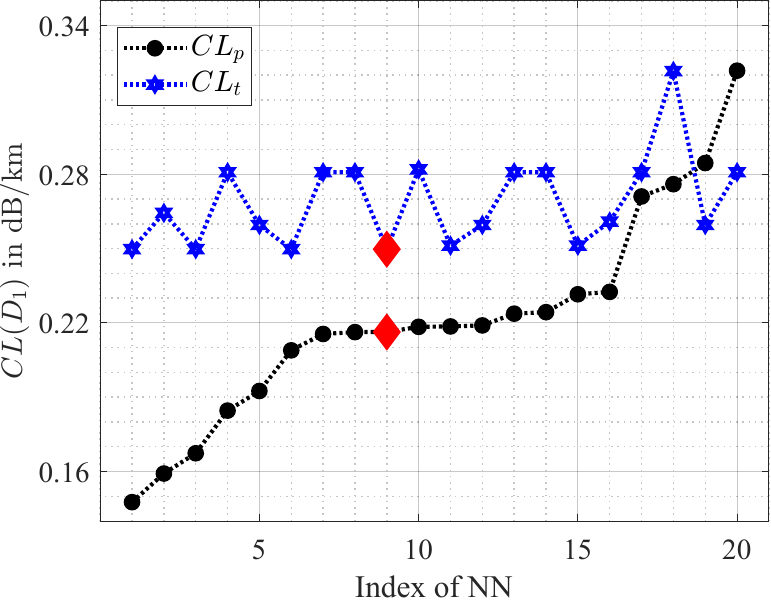}
\caption{Confirmed and predicted confinement loss of design~1 versus the \ac{NN} index with a data set of size $n$=1,819. } 
\label{fig_CL_vs_nmin}
\end{figure}

We sorted the results of the trials, with trial index of one for the lowest $CL_p(D_1)$, and 20 for the highest $CL_p(D_1)$.
In Fig.~\ref{fig_CL_vs_nmin} we plot  predicted $CL_p(D_1)$ and confirmed loss $CL_t(D_1)$ versus trial index. The large diamond marker indicates the trial that is reported in all previous results for $n=n_{min}$.

Consider the relative values of $CL_p(D_1)$ and $CL_t(D_1)$. The majority of cases have optimistic predictions, i.e., $CL_p(D_1)<CL_t(D_1)$. Recall that in Fig.~\ref{fig:Allset_CL_results} only one $n$ showed a pessimistic prediction, $n=n_{max}$. The pessimism may not be related to the size of $n$; perhaps it is simply that one of eight trials happened to be pessimistic.

\begin{figure}[!t]
\centering
{\includegraphics[width=0.47\textwidth]{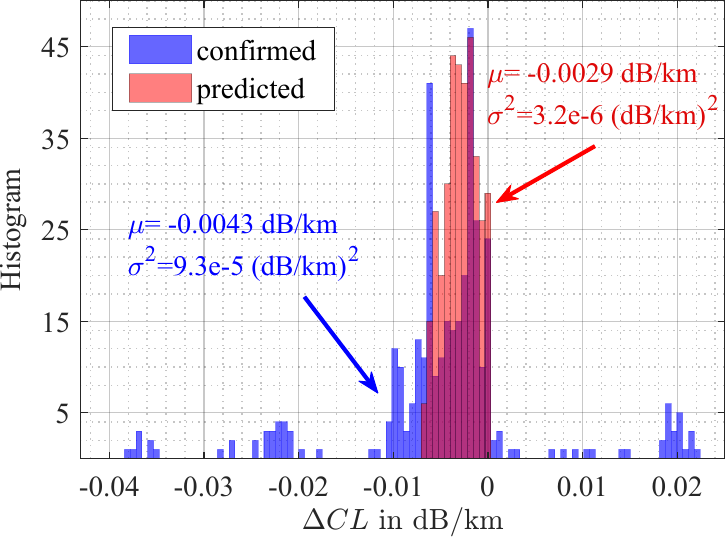}}
\caption{Histogram of~$\Delta CL$ for best 18 designs (relative to lowest predicted $CL(D_1)$ for the 20 tested \ac{NN} models of size $n$=1,819.}
\label{fig:Histogram_CLDeltaerror}
\end{figure}

Consider the range of predicted $CL_p(D_1)$ and confirmed $CL_t(D_1)$. The predicted $CL_p(D_1)$ spans 0.15~dB/km, but the confirmed loss $CL_t(D_1)$ only spans 0.05~dB/km. Whether optimistic or pessimistic, whatever design $D_1$ is found, its true loss indicates a significant improvement over the loss in the data set. We can also consider the distance between the curves (the error), which does not exceed 0.1~dB/km. This is the equivalent of 0.41 relative error per the definition (\ref{eq:relative_error}). This maximum among the twenty trials would fall in the tail of the estimated \ac{pdf} in Fig.~\ref{fig:regressor_error}. That \ac{pdf} was estimated from the sparse test portion of the data set; the much higher granularity of the search set makes it less surprising to fall in the tail of the distribution.

Finally, we consider the statistics of $CL_p(D_j)$ and  $CL_t(D_j)$ for $j\in$~(1,18), 360 designs in total. For each design,  we compute the error $\Delta CL$, defined by
\begin{equation}
    \Delta CL_j=CL(D_1)-CL(D_j).
\end{equation}

\noindent In Fig.~\ref{fig:Histogram_CLDeltaerror}, we plot the histograms of $\Delta CL$ of both predicted and confirmed loss. The histograms are skewed negative, meaning that, for both predicted and confirmed loss, the presumed best case $D_1$ is likely to truly be the best design. As expected from results in the preceding subsections (see Fig.~9), the predicted loss has a much more compact histogram than the true confirmed loss. The error in prediction is the source of the spread in the true loss. However, the span is still quite low, with a variance of {9.3e-5~$\text{(dB/km)}^2$.}

\section{Discussion}
\label{sec:discussion}

In the previous sections, we addressed the performance of the best identified designs at 1400~nm. The design yielding the red diamond marker in Fig.~\ref{fig_CL_vs_nmin} had 
$[ D_{core}  \; D_{cap} \; \alpha \; D_{nest} ]$ of $[31.0  \; 51.8 \; 0.22 \; 24.2]$, with diameters is $\mu$m. We study the performance of this design over the two antiresonance windows wavelengths.   

\begin{figure}[!t]
\centering
{\includegraphics[width=0.45\textwidth]{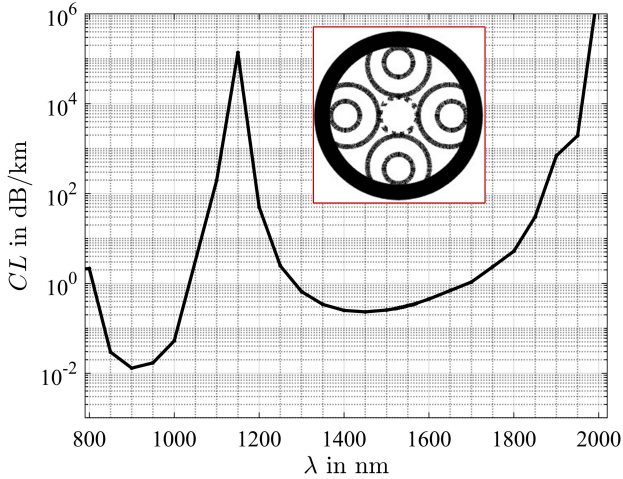}}
\caption{The confinement loss versus wavelength of the design in red diamond marker in Fig.~\ref{fig_CL_vs_nmin}. The inset is a scaled schematic of the fiber structure.}
\label{fig_best_design_performance}
\end{figure}

In Fig.~\ref{fig_best_design_performance}, we plot the fundamental mode confinement versus the wavelength for this design. The inset shows a scaled schematic of the selected \ac{NANF} structure. Over the C-band, the confinement loss never exceeds 0.4~dB/km. The confinement loss of the design does not exceed 1~dB/km over O-, C-, and L-bands. Despite training our \ac{NN} model on a data set with performance at 1400~nm, our model was able to identify a good design with stable performance over O-, C-, and L-bands.

\section{Conclusion}
\label{sec:conclusion}

In this study, we demonstrated a robust two-stage machine learning architecture for the rapid design and optimization of \ac{NANF} structures. By decoupling the identification of single-mode behavior from the prediction of confinement loss, our model effectively manages the vast dynamic range of optical performance metrics. The results demonstrate significant data efficiency; a data set of 1,819 designs proved sufficient to accurately navigate a search space of $14e6$ potential geometries. Furthermore, the model was able to identify optimized designs with a confirmed $CL$ of 0.25~dB/km, despite being trained exclusively on data with $CL\ge$1~dB/km. This suggests that the neural network captures the underlying physical trends of antiresonance guidance rather than merely interpolating within the training bounds. The stability of these findings across multiple data set realizations indicates that the proposed methodology offers a reliable, low-cost alternative to exhaustive finite element simulations, substantially accelerating the design cycle for hollow-core fibers.

 \bibliographystyle{IEEEtran}
 \bibliography{references}

\vfill

\end{document}